\documentclass[final,authoryear,5p,times,twocolumn]{elsarticle}
\usepackage{graphicx}
\usepackage{amssymb}
\usepackage{amsthm}
\usepackage{amsmath}
\journal{}

\begin{document}

\begin{frontmatter}

\title{A unifying view for performance measures in multi-class prediction}
\author{Giuseppe Jurman}
\ead{jurman@fbk.eu}
\author{Cesare Furlanello\corref{cor1}}
\address{Fondazione Bruno Kessler, Trento, Italy}
\cortext[cor1]{Corresponding author}
\ead{furlan@fbk.eu}

\begin{abstract}
In the last few years, many different performance measures have been introduced to overcome the weakness of the most natural metric, the Accuracy. 
Among them, Matthews Correlation Coefficient has recently gained popularity among researchers not only in machine learning but also in several application fields such as bioinformatics.
Nonetheless, further novel functions are being proposed in literature.
We show that Confusion Entropy, a recently introduced classifier performance measure for multi-class problems, has a strong (monotone) relation with the multi-class generalization of a classical metric, the Matthews Correlation Coefficient. 
Computational evidence in support of the claim is provided, together with an outline of the theoretical explanation.
\end{abstract}


\end{frontmatter}

\section{Introduction}
\label{sec:introduction}
One of the major task in machine learning is the comparison of classifiers' performance. 
This comparison can be carried out either by means of statistical tests \citep{demsar06statistical, garcia08extension} or using a performance measure as an indicator to derive similarities and differences.
For binary problems, a number of meaningful metrics are available and their properties are well understood.
On the other hand, the definition of performance measures in the context of multi-class classification is still an open research topic, although several functions have been proposed in the last few years: see \citep{sokolova09systematic, ferri09experimental} for two comparing reviews, \citep{felkin07comparing} for a discussion of the differences between the use of the same classifier on a binary and a multi-class task and \citep{diri08visualization} for an alternative graphical comparison approach. 
As an example, one of the most important measures for binary classifier, the Area Under the Curve (AUC) \citep{hanley82meaning,bradley97use} associated to the Receiver Operating Characteristic curve has no automatic extension to the multi-class case.
Although an agreed reasonably average-based build extension exists (presented in \citep{hand01simple}), several alternative formulations are being presented, either based on a multi-class ROC approximation \citep{everson06multiclass, landgrebe05neyman, landgrebe06simplified, landgrebe08efficient}) or by viewing the ROC as a surface whose volume (Volume Under the Surface, VUS) has to be computed (by exact integration or polynomial approximation) as in \citep{ferri03volume, vancalster08multiclass, li09generalization}.
Other measures are more naturally defined, starting from the accuracy (ACC, \textit{i.e.} the fraction of correctly predicted samples) and the similar Global Performance Index \citep{freitas07confusion,freitas07distance}), to the Matthews correlation coefficient (MCC).
This latter function was introduced in \citep{matthews75comparison} and it is also known as the $\phi$-coefficient, corresponding for a $2\times 2$ contingency table to the square root of the average $\chi^2$ statistic $\sqrt{\chi^2 / n}$.
MCC has recently attracted the attention of the machine learning community \citep{baldi00assessing} as one of the best method to summarize into a single value the confusion matrix of a binary classification task. 
Its use as one of the preferred classifier performance measure as increased since then, and for instance it has been chosen (together with AUC) as the elective metric in the US FDA-led initiative MAQC-II aimed at reaching consensus on the best practices for development and validation of predictive models based on microarray gene expression and genotyping data for personalized medicine \citep{maqc10maqcII}.
A generalization to the multi-class case was defined in \citep{gorodkin04comparing}, later used also for comparing network topologies \citep{supper07reconstructing,stokic09fast}.
Finally, another interesting set of measures that have a natural definition for multi-class confusion matrices consists of the functions derived from the concept of (information) entropy, first introduced by Shannon in his famous paper \citep{shannon48mathematical}.
Many measure have been defined in the classification framework based on the entropy function, from simpler ones such as the confusion matrix entropy \citep{vanson94method}, to more complex expressions as the transmitter information \citep{abramson63information} or the relative classifier information (RCI) \citep{sindhwani01information}.
A novel multi-class measure belonging to this set has been recently introduced under the name of Confusion Entropy (CEN) by Wei and colleagues in \citep{wei10novel, wei10evaluating}: in this work, the authors compare their measure to RCI and accuracy, and they prove CEN to be superior in discriminative power and precision to both alternatives in terms of two statistical indicator called degree of consistency and degree of discriminacy, defined in \citep{huang05using}.

In the present work we investigate the similarity between Confusion Entropy and Matthews correlation coefficient. 
In particular, we experimentally show that the two measures are strongly correlated, and their relation is globally monotone and locally almost linear.
Moreover, we provide a brief outline of the mathematical links between CEN and MCC.

\section{Confusion Entropy and Matthews Correlation Coefficient}
\label{sec:background}
Given a classification problem on $S$ samples $\mathcal{S}=\{s_i\colon 1\leq i\leq S\}$ and $N$ classes $\{1,\ldots, N\}$, define the two functions $\textrm{tc},\textrm{pc}\colon S\to \{1,\ldots,N\}$ indicating for each sample $s$ its true class $\textrm{tc}(s)$ and its predicted class $\textrm{pc}(s)$, respectively.
The corresponding confusion matrix is the square matrix $C\in\mathcal{M}(N\times N,\mathbb{N})$ whose $ij$-th entry $C_{ij}$ is the number of elements of true class $i$ that have been assigned to class $j$ by the classifier:
\begin{displaymath}
C_{ij} = | \{ s\in\mathcal{S}\colon \textrm{tc}(s)=i\;\textrm{and}\;\textrm{pc}(s)=j\} |\ .
\end{displaymath} 
The most natural performance measure is the accuracy, defined as the ratio of the correctly classified samples over all the samples:
\begin{displaymath}
\textrm{ACC} = \frac{ \displaystyle{\sum_{k=1}^N} C_{kk} }{S} = \frac{ \displaystyle{\sum_{k=1}^N} C_{kk} }{ \displaystyle{\sum_{i,j=1}^N} C_{ij} }\ .
\end{displaymath}

In information theory, the entropy $H$ associated to a random variable $X$ is the expected value of the self-information $I$ of $X$:
\begin{displaymath}
H(X) = \mathbb{E}(I(X))  = \sum_{x\in X} h_b(x) = - \sum_{x\in X} p(x) \log_b(p(x))\ ,
\end{displaymath}
where $p(x)$ is the probability mass function of $X$, with the position $h_b(x)=0$ for $p(x)=0$, motivated by the limit $\displaystyle{\lim_{x\to 0} x\log(x)=0}$.

The Confusion Entropy measure CEN for a confusion matrix $C$ is defined in \citep{wei10novel} as:
\begin{equation}
\label{eq:cen}
\textrm{CEN} = \sum_{j=1}^N P_j \sum_{\substack{k=1\\ k\not = j}}^{N} h_{2(N-1)}(P_{jk}^j)+h_{2(N-1)}(P_{kj}^j)\ ,
\end{equation}
where the misclassification probabilites $P$ are defined as the following ratios:
\begin{align*}
P_{ij}^j &= \frac{C_{ij}}{ \displaystyle{\sum_{k=1}^{N}} C_{jk}+C_{kj}} & P_{ii}^i &= 0 \\
P_{ij}^i &= \frac{C_{ij}}{ \displaystyle{\sum_{k=1}^{N}} C_{ik}+C_{ki}} & P_j &= \frac{\displaystyle{\sum_{k=1}^{N}} C_{jk}+C_{kj} }{2 \displaystyle{\sum_{k,l=1}^N} C_{kl}}\ .
\end{align*}
This measure ranges between $0$ (perfect classification) and $1$ for the extreme misclassification case $C_{ij} = (1-\delta_{ij})F$, for $F\in\mathbb{N}$ (this holds for $N>2$, while it is not true anymore for $N=2$, see Subsec.\ref{ssec:binary}).

Let $X,Y\in\mathcal{M}(S\times N,\mathbb{F}_2)$ be two matrices where $X_{sn}=1$ if the sample $s$ is predicted to of class $n$ ($\textrm{pc}(s)=n$) and $X_{sn}=0$ otherwise, and $Y_{sn}=1$ if sample $s$ belongs to class $n$ ($\textrm{tc}(s)=n$) and $0$ otherwise. 
Using Kronecker's delta function, the definition becomes:
\begin{displaymath}
X=\left(\delta_{\textrm{pc}(s),n}\right)_{sn}\quad
Y=\left(\delta_{\textrm{tc}(s),n}\right)_{sn}\ .
\end{displaymath}
Then the Matthews Correlation Coefficient MCC can be defined as the ratio:
\begin{displaymath}
\textrm{MCC} = \frac{\textrm{cov}(X,Y)}{\sqrt{{\textrm{cov}(X,X)}\cdot{\textrm{cov}(Y,Y)}}}\ ,
\end{displaymath}
where $\textrm{cov}(\cdot,\cdot)$ is the covariance function. In terms of the confusion matrix, the above equation can be written as:
\begin{equation}
\label{eq:mcc}
\textrm{MCC} = \frac{\displaystyle{\sum_{k,l,m=1}^N C_{kk}C_{ml} - C_{lk}C_{km}}}{
\sqrt{\displaystyle{\sum_{k=1}^N} \left(\displaystyle{\sum_{l=1}^N}C_{lk}\right) \left(\displaystyle{\sum_{\substack{f,g=1\\ f\not=k}}^N}C_{gf}\right)}
\sqrt{\displaystyle{\sum_{k=1}^N} \left(\displaystyle{\sum_{l=1}^N}C_{kl}\right) \left(\displaystyle{\sum_{\substack{f,g=1\\ f\not=k}}^N}C_{fg}\right)}
}
\end{equation}
MCC lives in the range $[-1,1]$, where $1$ is perfect classification, $-1$ is reached in the alternative extreme misclassification case of a confusion matrix with all zeros but in two symmetric entries $C_{\bar i,\bar j}$, $C_{\bar j,\bar i}$, and $0$ when the confusion matrix is all zeros but for one single column (all samples have been classified to be of a class $k$), or when all entries are equal $C_{ij}=K\in\mathbb{N}$. 
In this last case, the Confusion Entropy value is $\left( 1-\frac{1}{N}\right) \log_{2N-2} 2N$; when only a single column is not zero, the Confusion Entropy can assume many different values, depending on this column's entries.
Note that both measures are invariant for scalar multiplication of the whole confusion matrix.

CEN is indeed more discriminant than MCC in some situations, for instance when $\textrm{MCC}=0$ as mentioned above, or when the number of samples is relatively small and thus it more likely to have different confusion matrices with the same MCC and different CEN. 
This can be quantitatively assessed by using the degree of discrimination introduced in \citep{huang05using}: for two measures $f$ and $g$ on a domain $\Psi$, let $P=\{(a,b)\in\Psi\times\Psi\colon f(a)>f(b), g(a)=g(b)\}$ and $Q=\{(a,b)\in\Psi\times\Psi\colon f(a)=f(b), g(a)>g(b)\}$; then the degree of discriminancy for $f$ over $g$ is $|P|/|Q|$. 
For instance, in the 3-classes case with $2,4,3$ samples respectively, the degree of discriminancy of CEN over MCC is about 6. 
A similar behaviour happens for all the 12 small sample size cases on three classes listed in \citep[Tab. 6]{wei10novel}, ranging from 9 to 19 samples.
In the same paper \citep{huang05using}, another indicator for comparing distances is defined, the degree of consistency: for two measures $f$ and $ g$ on a domain $\Psi$, let $R=\{(a,b)\in\Psi\times\Psi\colon f(a)>f(b), g(a)>g(b)\}$ and $S=\{(a,b)\in\Psi\times\Psi\colon f(a)>f(b), g(a)<g(b)\}$; then the degree of consistency of $f$ and $g$ is $|R|/(|R|+|S|)$.

A quite different behaviour between the two measures can be highlighted in the following situation: consider the matrix $Z_A$ with all entries are equal but a non-diagonal one; because of the multiplicative invariance, we can set all entries to one but for the one in the leftmost lower corner: $(Z_A)_{ij} = 1+\delta_{(i,j),(N,1)}(A-1)$ for $A\geq 1$ a positive integer.
When $A$ grows bigger, more and more samples are misclassified: for instance, the corresponding accuracy reads $\textrm{ACC}(Z_A)=N / (N^2+A-1)$, thus decreasing towards zero for increasing $A$.

The MCC measure of this confusion matrix is 
\begin{displaymath}
\textrm{MCC}(Z_A) = - \frac{A-1}{(N-1)(N^2-2A-2)}\ ,
\end{displaymath}
which is a function monotonically decreasing for increasing values of $A$, with limit $-1/(N-1)$ for $A\to\infty$.
On the other hand, the Confusion Entropy for the same family of matrices is
\begin{displaymath}
\begin{split}
\textrm{CEN}(Z_A) &= \frac{1}{N^2+A-1}\left[ (N-2)(N-1)\log_{2N-2}(2N)\right.\\
&\quad \left. +(2N+A-3)\log_{2N-2}(2N+A-1)-A\log_{2N-2}(A)\right]\ ,
\end{split}
\end{displaymath}
which is a decreasing function of increasing $A$, asymptotically moving towards zero, i.e., the minimal entropy case.
Thus in this case, the behaviour of the Confusion Entropy is the opposite than the one of more classical measures such as MCC and accuracy.

Analogously for the case of (perfectly) random classification on a unbalanced problem: because of the multiplicative invariance of the measures, we can assume that the confusion matrix for this case has all entries equal to one but for the last row, whose entries are all $A$, for $A\geq 1$. 
In this case, the Confusion Entropy is
\begin{displaymath}
\begin{split}
\textrm{CEN} &= \frac{N-1}{2N(N+A-1)}\left[ (2N+A-3)\log_{2N-2}(2N+A-1)\right. \\
&\quad \left. - 2A\log_{2N-2}A + (A+1)\log_{2N-2}(N+NA+A-1)\right]\ ,
\end{split}
\end{displaymath}
which is a decreasing function for growing $A$ whose limit for $A\to\infty$ is $\frac{N-1}{2N}\log_{2N-2}N+1$ (as a function of $N$, this limit is an increasing function asymptotically growing towards $1/2$).

One of the main features of the MCC measure is the fact that MCC=0 identifies all those case where random classification (i.e., no learning) happens: this is lost in the case of CEN, due to its greater discriminant power - there is no unique value associated to the wide spectrum of random classification.

Consider now the confusion matrix $B$ of dimension $N$ where $B_{ji}=F+(T-F)\delta_{ij}$, i.e. all entries have value $F$ but in the diagonal whose values are all $T$, for $T$, $F$ two integers.
In this case, 
\begin{displaymath}
\begin{split}
\textrm{MCC} &= \frac{ T^2 +(N-2) TF -(N-1)F^2}{ [T+(N-1)F]^2}\\
& \\
\textrm{CEN} &= \frac{(N-1)F}{T+(N-1)F}\log_{2N-2} \frac{2[T+(N-1)F]}{F}\ ,\\
\end{split}
\end{displaymath}
and thus
\begin{displaymath}
\textrm{CEN} = (1-\textrm{MCC})\left( 1+\log_{2N-2} \frac{T+(N-1)F}{(N-1)F} \right)\left(1-\frac{1}{N}\right)\ .
\end{displaymath}

This identity can be relaxed to the following generalization, which is a slight underestimate of the true CEN value:
\begin{equation}
\label{eq:tmcc}
\begin{split}
\textrm{CEN} &\simeq \frac{1}{k}\cdot (1-\textrm{MCC})\left( 1+\log_{2N-2} \frac{ \displaystyle{\sum_{i,j=1}^N C_{ij}} }{ \displaystyle{\sum_{\substack{i,j=1\\ i\not=j}}^N C_{ij}} } \right)\left(1-\frac{1}{N}\right)\\
&\simeq \frac{1}{k}\cdot (1-\textrm{MCC})\left( 1-\log_{2N-2} (1-\textrm{ACC})\right)\left(1-\frac{1}{N}\right)\\
\end{split}
\end{equation}
where both sides are zero when $\textrm{MCC}=\textrm{ACC}=1$, and $k=1.012\cdot\left(1+\frac{0.18924}{\log(N)}-\frac{0.06694}{\log^2(N)}\right)$.
For simplicity sake, we call the right member of Eq. \ref{eq:tmcc} transformed MMC, tMCC for short.

\begin{figure*}[t]
\includegraphics[angle=-90,width=\columnwidth]{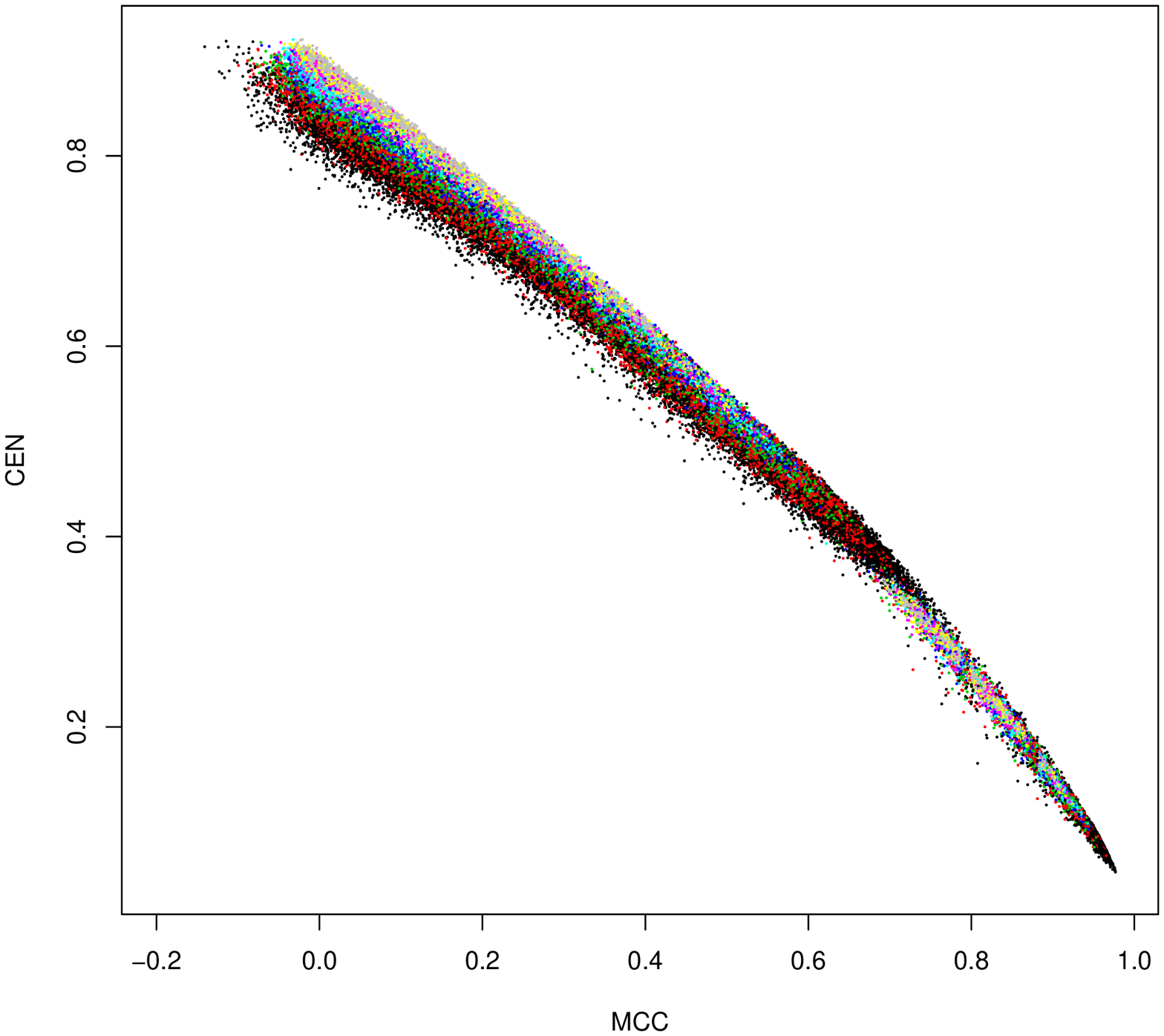}
\includegraphics[angle=-90,width=\columnwidth]{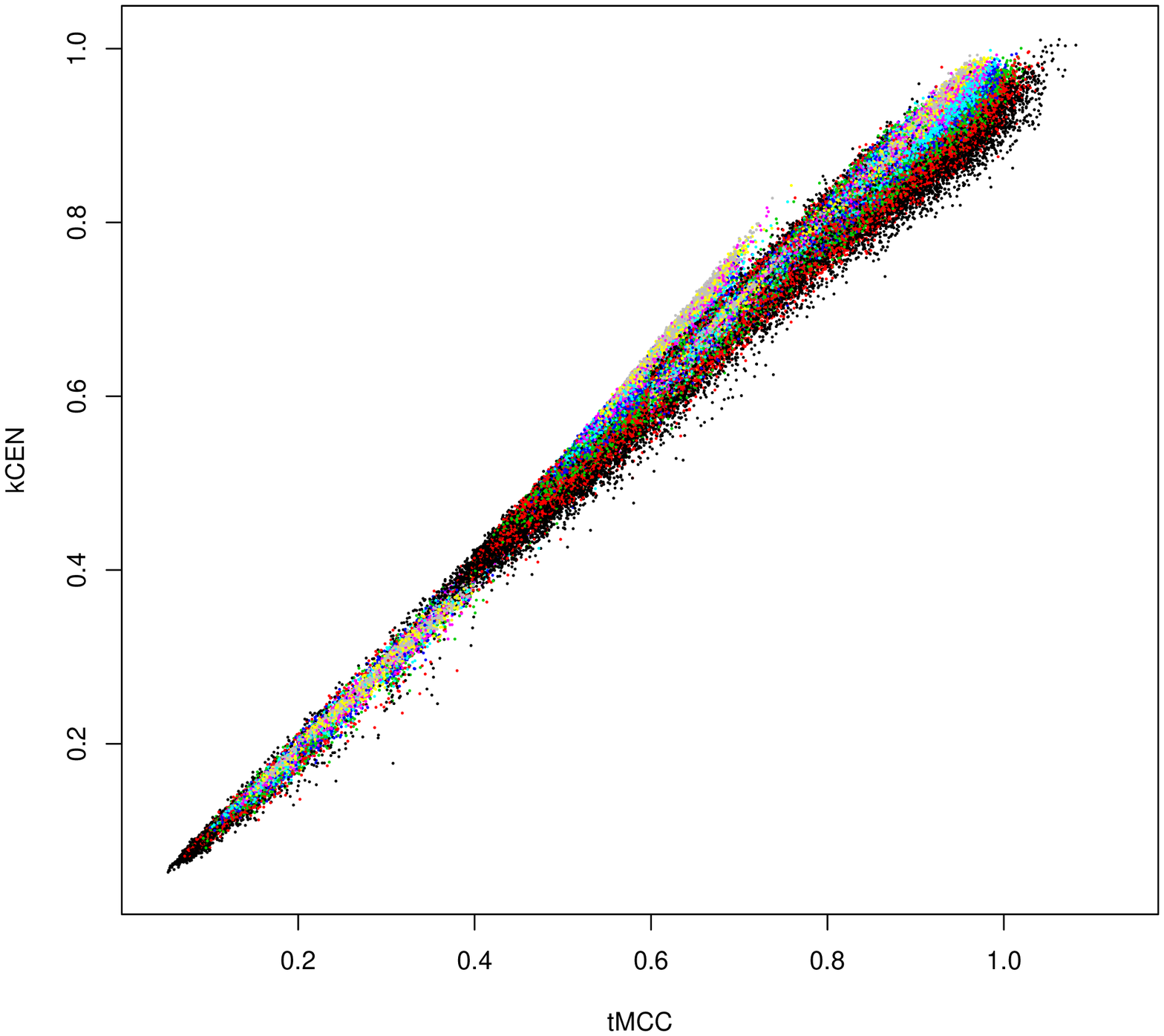}
\caption{Plot of CEN versus MCC (left) and $k\cdot$CEN versus tMCC (right) for 200.000 random confusion matrices. Each dot represents a confusion matrix, and the color indicates the matrix dimension.}
\label{fig:exp1}
\end{figure*}

To show that the relation in Eq. \ref{eq:tmcc} is valid in a wide range of situations, an experiment has been performed, whose result is graphically reported in Fig. \ref{fig:exp1}, 
In details, 200.000 confusion matrices in dimensions ranging from 3 to 30 have been generated with the following setup: the number correctly classified elements (i.e., the diagonal elements) for each class has been (uniformly) randomly chosen between 1 and 1000, while each non-diagonal entry has been chosen as a random integer between 1 and $\lfloor 1000\rho_i\rfloor$, where the ratio $\rho_i$ for the $i$-th matrix $M_i$ was extracted from the uniform distribution in the range $[0.01,1]$.
The correlation between tMCC and $k\cdot$CEN is 0.9941477 and the degree of consistency is $1-10^{-7}$ (the degree of discriminancy is undefined since no ties occurred).
In particular, the average ratio between tMMC and $k\cdot$CEN is 1.000508, with 95\% bootstrap Student confidence interval $(1.000328, 1.000711)$.

\subsection{The binary case}
\label{ssec:binary}
In the binary case of two classes positive ($P$) and negative ($N$), the confusion matrix becomes $\left(\begin{smallmatrix} \textrm{TP} & \textrm{FN} \\ \textrm{FP} & \textrm{TN}\end{smallmatrix}\right)$, where $T$ and $F$ stands for true and false respectively.

In this setup, the Matthews correlation coefficient has the following shape:
\begin{displaymath}
\textrm{MCC} = \frac{\textrm{TP}\cdot\textrm{TN}-\textrm{FP}\cdot\textrm{FN}}{\sqrt{\left(\textrm{TP}+\textrm{FP}\right)\left(\textrm{TP}+\textrm{FN}\right)\left(\textrm{TN}+\textrm{FP}\right)\left(\textrm{TN}+\textrm{FN}\right)}}\ .
\end{displaymath}

Similarly, the Confusion Entropy can be written as:
\begin{displaymath}
\begin{split}
\textrm{CEN} &= \frac
{(\textrm{FN}+\textrm{FP})\log_2((\textrm{TP}+\textrm{TN}+\textrm{FP}+\textrm{FN})^2-(\textrm{TP}-\textrm{TN})^2)}
{2(\textrm{TP}+\textrm{TN}+\textrm{FP}+\textrm{FN})} \\
&\quad  - \frac
{\textrm{FN}\log_2\textrm{FN}+\textrm{FP}\log_2\textrm{FP}}
{\textrm{TP}+\textrm{TN}+\textrm{FP}+\textrm{FN}} \ .
\end{split}
\end{displaymath}
Note that in the case $\textrm{TP}=\textrm{TN}=T$ and $\textrm{FP}=\textrm{FN}=F$, the Confusion Entropy reads 
\begin{displaymath}
\textrm{CEN} = \frac{F}{T+F}\log_2\frac{2(T+F)}{F}\ ,
\end{displaymath}
which is bigger than $1$ when the ratio $T / F$ is smaller than 1. 
This means that all the confusion matrices $\left( \begin{smallmatrix} T & F\\ F & T \end{smallmatrix}\right)$ with $0<T<F$ have a confusion entropy larger than 1, attained for the totally misclassified case $T=0$. 
Such behaviour makes CEN unusable as a classifier performance measure in the binary case.

\section{Conclusions}
\label{sec:conclusions}
Accuracy, Matthews Correlation Coefficient and Confusion Entropy are three crucial performance measures for evaluating the outcome of a classification task, both on binary and multi-class problems (the fourth one is Area Under the Curve, whenever a ROC curve can be drawn).
Although they show a mutual consistent behaviour, each of them is better tailored to deal with different situations.

Accuracy is by far the simplest one, and its role is to convey a first rough estimate of the classifier goodness. Its use is widespread among the scientific literature, but it suffers from several caveats, the most relevant being the inability to cope with unbalanced classes and thus the impossibility of distinguish among different kinds of misclassifications.

Confusion Entropy, on the other hand, is probably the finest measure and it shows an extremely high level of discriminancy even between very similar confusion matrices. However, this feature is not always welcomed, because it makes the interpretation of its value quite harder, expecially when considering situations that are naturally very similar (e.g, all the cases with MCC=0). Moreover, CEN may 
show erratic behaviour in the binary case.

In this spirit, the Matthews Correlation Coefficient is a good compromise between reaching a reasonable discriminancy degree among different cases, and the need for the practitioner of a easily interpretable value expressing the type of misclassification associated to the chosen classifier on the given task. We showed here that there is a strong linear relation between CEN and a logarithmic function of MCC regardless of the dimension of the considered problem. Furthermore, MCC behaviour is totally consistent also for the binary case. 

This given, we can suggest MCC as the best off-the-shelf evaluating tool for general purpose tasks, while more subtle measures such as CEN should be reserved for specific topic where more refined discrimination is crucial.

\bibliographystyle{model5-names}
\bibliography{jurman10unifying}

\begin{thebibliography}{31}
\expandafter\ifx\csname natexlab\endcsname\relax\def\natexlab#1{#1}\fi
\providecommand{\bibinfo}[2]{#2}
\ifx\xfnm\relax \def\xfnm[#1]{\unskip,\space#1}\fi
\bibitem[{Abramson(1963)}]{abramson63information}
\bibinfo{author}{Abramson, N.} (\bibinfo{year}{1963}).
\newblock {\it \bibinfo{title}{Information theory and coding}\/}.
\newblock \bibinfo{publisher}{McGraw-Hill}.
\bibitem[{Baldi et~al.(2000)Baldi, Brunak, Chauvin, Andersen \&
  Nielsen}]{baldi00assessing}
\bibinfo{author}{Baldi, P.}, \bibinfo{author}{Brunak, S.},
  \bibinfo{author}{Chauvin, Y.}, \bibinfo{author}{Andersen, C.}, \&
  \bibinfo{author}{Nielsen, H.} (\bibinfo{year}{2000}).
\newblock \bibinfo{title}{{Assessing the accuracy of prediction algorithms for
  classification: an overview}}.
\newblock {\it \bibinfo{journal}{Bioinformatics}\/},  {\it
  \bibinfo{volume}{16}\/}, \bibinfo{pages}{412--424}.
\bibitem[{Bradley(1997)}]{bradley97use}
\bibinfo{author}{Bradley, A.} (\bibinfo{year}{1997}).
\newblock \bibinfo{title}{{The use of the area under the ROC curve in the
  evaluation of machine learning algorithms}}.
\newblock {\it \bibinfo{journal}{Pattern Recognition}\/},  {\it
  \bibinfo{volume}{30}\/}, \bibinfo{pages}{1145--1159}.
\bibitem[{Dem\v{s}ar(2006)}]{demsar06statistical}
\bibinfo{author}{Dem\v{s}ar, J.} (\bibinfo{year}{2006}).
\newblock \bibinfo{title}{{Statistical Comparisons of Classifiers over Multiple
  Data Sets}}.
\newblock {\it \bibinfo{journal}{Journal of Machine Learning Research}\/},
  {\it \bibinfo{volume}{7}\/}, \bibinfo{pages}{1--30}.
\bibitem[{Diri \& Albayrak(2008)}]{diri08visualization}
\bibinfo{author}{Diri, B.}, \& \bibinfo{author}{Albayrak, S.}
  (\bibinfo{year}{2008}).
\newblock \bibinfo{title}{{Visualization and analysis of classifiers
  performance in multi-class medical data}}.
\newblock {\it \bibinfo{journal}{Expert Systems with Applications}\/},  {\it
  \bibinfo{volume}{34}\/}, \bibinfo{pages}{628--634}.
\bibitem[{Everson \& Fieldsend(2006)}]{everson06multiclass}
\bibinfo{author}{Everson, R.}, \& \bibinfo{author}{Fieldsend, J.}
  (\bibinfo{year}{2006}).
\newblock \bibinfo{title}{{Multi-class ROC analysis from a multi-objective
  optimisation perspective}}.
\newblock {\it \bibinfo{journal}{Pattern Recognition Letters}\/},  {\it
  \bibinfo{volume}{27}\/}, \bibinfo{pages}{918--927}.
\bibitem[{Felkin(2007)}]{felkin07comparing}
\bibinfo{author}{Felkin, M.} (\bibinfo{year}{2007}).
\newblock \bibinfo{title}{{Comparing Classification Results between N-ary and
  Binary Problems}}.
\newblock In {\it \bibinfo{booktitle}{{Studies in Computational
  Intelligence}}\/} (pp. \bibinfo{pages}{277--301}).
\newblock \bibinfo{publisher}{Springer-Verlag} volume~\bibinfo{volume}{43}.
\bibitem[{Ferri et~al.(2009)Ferri, Hern\'{a}ndez-Orallo \&
  Modroiu}]{ferri09experimental}
\bibinfo{author}{Ferri, C.}, \bibinfo{author}{Hern\'{a}ndez-Orallo, J.}, \&
  \bibinfo{author}{Modroiu, R.} (\bibinfo{year}{2009}).
\newblock \bibinfo{title}{An experimental comparison of performance measures
  for classification}.
\newblock {\it \bibinfo{journal}{Pattern Recognition Letters}\/},  {\it
  \bibinfo{volume}{30}\/}, \bibinfo{pages}{27--38}.
\bibitem[{Ferri et~al.(2003)Ferri, Hern\'{a}ndez-Orallo \&
  Salido}]{ferri03volume}
\bibinfo{author}{Ferri, C.}, \bibinfo{author}{Hern\'{a}ndez-Orallo, J.}, \&
  \bibinfo{author}{Salido, M.} (\bibinfo{year}{2003}).
\newblock \bibinfo{title}{{Volume under the ROC surface for multi-class
  problems}}.
\newblock In {\it \bibinfo{booktitle}{In Proc. of 14th European Conference on
  Machine Learning}\/} (pp. \bibinfo{pages}{108--120}).
\newblock \bibinfo{publisher}{Springer-Verlag}.
\bibitem[{Freitas et~al.(2007{\natexlab{a}})Freitas, De~Carvalho, Oliveira~Jr.,
  Aires \& Sabourin}]{freitas07confusion}
\bibinfo{author}{Freitas, C.}, \bibinfo{author}{De~Carvalho, J.},
  \bibinfo{author}{Oliveira~Jr., J.}, \bibinfo{author}{Aires, S.}, \&
  \bibinfo{author}{Sabourin, R.} (\bibinfo{year}{2007}{\natexlab{a}}).
\newblock \bibinfo{title}{Confusion matrix disagreement for multiple
  classifiers}.
\newblock In \bibinfo{editor}{L.~Rueda}, \bibinfo{editor}{D.~Mery}, \&
  \bibinfo{editor}{J.~Kittler} (Eds.), {\it \bibinfo{booktitle}{Proceedings of
  12th Iberoamerican Congress on Pattern Recognition, CIARP 2007, LNCS 4756}\/}
  (pp. \bibinfo{pages}{387--396}).
\newblock \bibinfo{publisher}{Springer-Verlag}.
\bibitem[{Freitas et~al.(2007{\natexlab{b}})Freitas, De~Carvalho, Oliveira~Jr.,
  Aires \& Sabourin}]{freitas07distance}
\bibinfo{author}{Freitas, C.}, \bibinfo{author}{De~Carvalho, J.},
  \bibinfo{author}{Oliveira~Jr., J.}, \bibinfo{author}{Aires, S.}, \&
  \bibinfo{author}{Sabourin, R.} (\bibinfo{year}{2007}{\natexlab{b}}).
\newblock \bibinfo{title}{{Distance-based Disagreement Classifiers
  Combination}}.
\newblock In {\it \bibinfo{booktitle}{Proceedings of the International Joint
  Conference on Neural Networks, IJCNN 2007}\/} (pp.
  \bibinfo{pages}{2729--2733}).
\newblock \bibinfo{publisher}{IEEE}.
\bibitem[{Garc\'{\i}a \& Herrera(2008)}]{garcia08extension}
\bibinfo{author}{Garc\'{\i}a, S.}, \& \bibinfo{author}{Herrera, F.}
  (\bibinfo{year}{2008}).
\newblock \bibinfo{title}{{An Extension on ''Statistical Comparisons of
  Classifiers over Multiple Data Sets'' for all Pairwise Comparisons}}.
\newblock {\it \bibinfo{journal}{Journal of Machine Learning Research}\/},
  {\it \bibinfo{volume}{9}\/}, \bibinfo{pages}{2677--2694}.
\bibitem[{Gorodkin(2004)}]{gorodkin04comparing}
\bibinfo{author}{Gorodkin, J.} (\bibinfo{year}{2004}).
\newblock \bibinfo{title}{{Comparing two K-category assignments by a K-category
  correlation coefficient}}.
\newblock {\it \bibinfo{journal}{Computational Biology and Chemistry}\/},  {\it
  \bibinfo{volume}{28}\/}, \bibinfo{pages}{367--374}.
\bibitem[{Hand \& Till(2001)}]{hand01simple}
\bibinfo{author}{Hand, D.}, \& \bibinfo{author}{Till, R.}
  (\bibinfo{year}{2001}).
\newblock \bibinfo{title}{{A Simple Generalisation of the Area Under the ROC
  Curve for Multiple Class Classification Problems}}.
\newblock {\it \bibinfo{journal}{Machine Learning}\/},  {\it
  \bibinfo{volume}{45}\/}, \bibinfo{pages}{171--186}.
\bibitem[{Hanley \& McNeil(1982)}]{hanley82meaning}
\bibinfo{author}{Hanley, J.}, \& \bibinfo{author}{McNeil, B.}
  (\bibinfo{year}{1982}).
\newblock \bibinfo{title}{{The meaning and use of the area under a receiver
  operating characteristic (ROC) curve}}.
\newblock {\it \bibinfo{journal}{Radiology}\/},  {\it \bibinfo{volume}{143}\/},
  \bibinfo{pages}{29--36}.
\bibitem[{Huang \& Ling(2005)}]{huang05using}
\bibinfo{author}{Huang, J.}, \& \bibinfo{author}{Ling, C.}
  (\bibinfo{year}{2005}).
\newblock \bibinfo{title}{{Using AUC and Accuracy in Evaluating Learning
  Algorithms}}.
\newblock {\it \bibinfo{journal}{IEEE Transactions on Knowledge and Data
  Engineering}\/},  {\it \bibinfo{volume}{17}\/}, \bibinfo{pages}{299--310}.
\bibitem[{Landgrebe \& Duin(2005)}]{landgrebe05neyman}
\bibinfo{author}{Landgrebe, T.}, \& \bibinfo{author}{Duin, R.}
  (\bibinfo{year}{2005}).
\newblock \bibinfo{title}{{On Neyman-Pearson optimisation for multiclass
  classifiers}}.
\newblock In {\it \bibinfo{booktitle}{Proc. 16th Annual Symposium of the
  Pattern Recognition Assoc. of South Africa}\/}.
\newblock \bibinfo{publisher}{PRASA}.
\bibitem[{Landgrebe \& Duin(2006)}]{landgrebe06simplified}
\bibinfo{author}{Landgrebe, T.}, \& \bibinfo{author}{Duin, R.}
  (\bibinfo{year}{2006}).
\newblock \bibinfo{title}{{A simplified extension of the Area under the ROC to
  the multiclass domain}}.
\newblock In {\it \bibinfo{booktitle}{Proc. 17th Annual Symposium of the
  Pattern Recognition Assoc. of South Africa}\/} (pp.
  \bibinfo{pages}{241--245}).
\newblock \bibinfo{publisher}{PRASA}.
\bibitem[{Landgrebe \& Duin(2008)}]{landgrebe08efficient}
\bibinfo{author}{Landgrebe, T.}, \& \bibinfo{author}{Duin, R.}
  (\bibinfo{year}{2008}).
\newblock \bibinfo{title}{{Efficient multiclass ROC approximation by
  decomposition via confusion matrix perturbation analysis}}.
\newblock {\it \bibinfo{journal}{IEEE Transactions Pattern Analysis Machine
  Intelligence}\/},  {\it \bibinfo{volume}{30}\/}, \bibinfo{pages}{810--822}.
\bibitem[{Li(2009)}]{li09generalization}
\bibinfo{author}{Li, Y.} (\bibinfo{year}{2009}).
\newblock {\it \bibinfo{title}{{A generalization of AUC to an ordered
  multi-class diagnosis and application to longitudinal data analysis on
  intellectual outcome in pediatric brain-tumor patients}}\/}.
\newblock Ph.D. thesis College of Arts and Sciences, Georgia State University.
\bibitem[{Matthews(1975)}]{matthews75comparison}
\bibinfo{author}{Matthews, B.} (\bibinfo{year}{1975}).
\newblock \bibinfo{title}{{Comparison of the predicted and observed secondary
  structure of T4 phage lysozyme}}.
\newblock {\it \bibinfo{journal}{Biochimica et Biophysica Acta - Protein
  Structure}\/},  {\it \bibinfo{volume}{405}\/}, \bibinfo{pages}{442--451}.
\bibitem[{Shannon(1948)}]{shannon48mathematical}
\bibinfo{author}{Shannon, C.} (\bibinfo{year}{1948}).
\newblock \bibinfo{title}{{A Mathematical Theory of Communication}}.
\newblock {\it \bibinfo{journal}{The Bell System Technical Journal}\/},  {\it
  \bibinfo{volume}{27}\/}, \bibinfo{pages}{379--423, 623--656}.
\bibitem[{Sindhwani et~al.(2001)Sindhwani, Bhattacharge \&
  Rakshit}]{sindhwani01information}
\bibinfo{author}{Sindhwani, V.}, \bibinfo{author}{Bhattacharge, P.}, \&
  \bibinfo{author}{Rakshit, S.} (\bibinfo{year}{2001}).
\newblock \bibinfo{title}{{Information theoretic feature crediting in
  multiclass Support Vector Machines}}.
\newblock In \bibinfo{editor}{R.~Grossman}, \& \bibinfo{editor}{V.~Kumar}
  (Eds.), {\it \bibinfo{booktitle}{Proc. First SIAM International Conference on
  Data Mining, ICDM01}\/} (pp. \bibinfo{pages}{1--18}).
\newblock \bibinfo{publisher}{SIAM}.
\bibitem[{Sokolova \& Lapalme(2009)}]{sokolova09systematic}
\bibinfo{author}{Sokolova, M.}, \& \bibinfo{author}{Lapalme, G.}
  (\bibinfo{year}{2009}).
\newblock \bibinfo{title}{A systematic analysis of performance measures for
  classification tasks}.
\newblock {\it \bibinfo{journal}{Information Processing and Management}\/},
  {\it \bibinfo{volume}{45}\/}, \bibinfo{pages}{427--437}.
\bibitem[{van Son(1994)}]{vanson94method}
\bibinfo{author}{van Son, R.} (\bibinfo{year}{1994}).
\newblock {\it \bibinfo{title}{{A method to quantify the error distribution in
  confusion matrices}}\/}.
\newblock \bibinfo{type}{Technical Report} \bibinfo{number}{IFA Proceedings 18}
  Institute of Phonetic Sciences, University of Amsterdam.
\bibitem[{Stokic et~al.(2009)Stokic, Hanel \& Thurner}]{stokic09fast}
\bibinfo{author}{Stokic, D.}, \bibinfo{author}{Hanel, R.}, \&
  \bibinfo{author}{Thurner, S.} (\bibinfo{year}{2009}).
\newblock \bibinfo{title}{A fast and efficient gene-network reconstruction
  method from multiple over-expression experiments}.
\newblock {\it \bibinfo{journal}{BMC Bioinformatics}\/},  {\it
  \bibinfo{volume}{10}\/}, \bibinfo{pages}{253}.
\bibitem[{Supper et~al.(2007)Supper, Spieth \& Zell}]{supper07reconstructing}
\bibinfo{author}{Supper, J.}, \bibinfo{author}{Spieth, C.}, \&
  \bibinfo{author}{Zell, A.} (\bibinfo{year}{2007}).
\newblock \bibinfo{title}{{Reconstructing Linear Gene Regulatory Networks}}.
\newblock In \bibinfo{editor}{E.~Marchiori}, \bibinfo{editor}{J.~Moore}, \&
  \bibinfo{editor}{J.~Rajapakse} (Eds.), {\it \bibinfo{booktitle}{Proceedings
  of the 5th European Conference on Evolutionary Computation, Machine Learning
  and Data Mining in Bioinformatics, EvoBIO2007, LNCS 4447}\/} (pp.
  \bibinfo{pages}{270--279}).
\newblock \bibinfo{publisher}{Springer-Verlag}.
\bibitem[{{The MicroArray Quality Control (MAQC)
  Consortium}(2010)}]{maqc10maqcII}
\bibinfo{author}{{The MicroArray Quality Control (MAQC) Consortium}}
  (\bibinfo{year}{2010}).
\newblock \bibinfo{title}{{The MAQC-II Project: A comprehensive study of common
  practices for the development and validation of microarray-based predictive
  models}}.
\newblock {\it \bibinfo{journal}{Nature Biotechnology}\/},  {\it
  \bibinfo{volume}{28}\/}, \bibinfo{pages}{827--838}.
\bibitem[{Van~Calster et~al.(2008)Van~Calster, Van~Belle, Condous, Bourne,
  Timmerman \& Van~Huffel}]{vancalster08multiclass}
\bibinfo{author}{Van~Calster, B.}, \bibinfo{author}{Van~Belle, V.},
  \bibinfo{author}{Condous, G.}, \bibinfo{author}{Bourne, T.},
  \bibinfo{author}{Timmerman, D.}, \& \bibinfo{author}{Van~Huffel, S.}
  (\bibinfo{year}{2008}).
\newblock \bibinfo{title}{{Multi-class AUC metrics and weighted alternatives}}.
\newblock In {\it \bibinfo{booktitle}{Proc. 2008 International Joint Conference
  on Neural Networks, IJCNN08}\/} (pp. \bibinfo{pages}{1390--1396}).
\newblock \bibinfo{publisher}{IEEE}.
\bibitem[{Wei et~al.(2010{\natexlab{a}})Wei, Yuan, Hu \& Wang}]{wei10novel}
\bibinfo{author}{Wei, J.-M.}, \bibinfo{author}{Yuan, X.-J.},
  \bibinfo{author}{Hu, Q.-H.}, \& \bibinfo{author}{Wang, S.-Q.}
  (\bibinfo{year}{2010}{\natexlab{a}}).
\newblock \bibinfo{title}{A novel measure for evaluating classifiers}.
\newblock {\it \bibinfo{journal}{Expert Systems with Applications}\/},  {\it
  \bibinfo{volume}{37}\/}, \bibinfo{pages}{3799--3809}.
\bibitem[{Wei et~al.(2010{\natexlab{b}})Wei, Yuan, Yang \&
  Wang}]{wei10evaluating}
\bibinfo{author}{Wei, J.-M.}, \bibinfo{author}{Yuan, X.-J.},
  \bibinfo{author}{Yang, T.}, \& \bibinfo{author}{Wang, S.-Q.}
  (\bibinfo{year}{2010}{\natexlab{b}}).
\newblock \bibinfo{title}{{Evaluating Classifiers by Confusion Entropy}}.
\newblock {\it \bibinfo{journal}{Information Processing \& Management}\/},
  {\it \bibinfo{volume}{Submitted}\/}.

\end{thebibliography}
\end{document}